\documentclass[a4paper]{article}

\usepackage{INTERSPEECH2018}
\usepackage{multirow}

\title{Visual Speech Enhancement}
\name{Aviv Gabbay, Asaph Shamir, Shmuel Peleg}

\address{
School of Computer Science and Engineering\\
The Hebrew University of Jerusalem\\
Jerusalem, Israel
}
\email{}

\begin{document}

\maketitle

\begin{abstract}

When video is shot in noisy environment, the voice of a speaker seen in the video can be enhanced using the visible mouth movements, reducing background noise.
While most existing methods use audio-only inputs, improved performance is obtained with our visual speech enhancement, based on an audio-visual neural network.

We include in the training data videos to which we added the voice of the target speaker as background noise. Since the audio input is not sufficient to separate the voice of a speaker from his own voice, the trained model better exploits the visual input and generalizes well to different noise types.

The proposed model outperforms prior audio visual methods on two public lipreading datasets. It is also the first to be demonstrated on a dataset not designed for lipreading, such as the weekly addresses of Barack Obama.

\end{abstract}


\section{Introduction}
\label{sec:intro}

Speech enhancement aims to improve speech quality and intelligibility when audio is recorded in noisy environments. Applications include telephone conversations, video conferences, TV reporting and more. Speech enhancement can also be used in hearing aids \cite{yang2005spectral}, speech recognition, and speaker identification \cite{yu2008minimum, maas2012recurrent}.
Speech enhancement has been the subject of extensive research \cite{Bronkhorst2000, ephraim1984speech, Loizou2013}, and has recently benefited from advancements in machine lip reading \cite{chung2016lip,assael2016lipnet} and speech reading \cite{cornu2017,ephrat2017improved,vid2speech}.

We propose an audio-visual end-to-end neural network model for separating the voice of a visible speaker from background noise. Once the model is trained on a specific speaker, it can be used to enhance the voice of this speaker. We assume a video showing the face of the target speaker is available along with the noisy soundtrack, and use the visible mouth movements to isolate the desired voice from the background noise. 

While the idea of training a deep neural network to differentiate between the unique speech or auditory characteristics of different sources can be very effective in several cases, the performance is limited by the variance of the sources, as shown in \cite{isik2016single,chen2017single}. We show that using the visual information leads to significant improvement in the enhancement performance in different scenarios. In order to cover cases where the target and background speech can not be totally separated using the audio information alone, we add to the training data videos with synthetic background noise taken from the voice of the target speaker. With such videos in the training data, the trained model better exploits the visual input and generalizes well to different noise types.

We evaluate the performance of our model in different speech enhancement experiments.
First, we show better performance compared to prior art on two common audio-visual datasets: GRID corpus \cite{gridcorpus} and TCD-TIMIT \cite{harte2015tcd}, both designed for audio-visual speech recognition and lip reading.
We also demonstrate speech enhancement on public weekly addresses of Barack Obama.

\subsection{Related work}

Traditional speech enhancement methods include spectral restoration \cite{scalart1996speech,ephraim1984speech}, Wiener filtering \cite{lim1978all} and statistical model-based methods \cite{ephraim1992statistical}. Recently, deep neural networks have been adopted for speech enhancement \cite{luspeech,parveen2004speech,isik2016single}, generally outperforming the traditional methods \cite{kolbaek2017speech}.

\subsubsection{Audio-only deep learning based speech enhancement}
Previous methods for single-channel speech enhancement mostly use audio only input. Lu \emph{et al.} \cite{luspeech} train a deep auto-encoder for denoising the speech signal. Their model predicts a mel-scale spectrogram representing the clean speech. Pascual \emph{et al.} \cite{pascual2017segan} use generative adversarial networks and operate at the waveform level. Separating mixtures of several people speaking simultaneously has also become possible by training a deep neural network to differentiate between the unique speech characteristics of different sources e.g. spectral bands, pitches and chirps, as shown in \cite{isik2016single,chen2017single}.
Despite their decent overall performance, audio-only approaches achieve lower performance when separating similar human voices, as commonly observed in same-gender mixtures \cite{isik2016single}.

\subsubsection{Visually-derived speech and sound generation}
Different approaches exist for generation of intelligible speech from silent video frames of a speaker \cite{vid2speech, ephrat2017improved, cornu2017}. In \cite{ephrat2017improved}, Ephrat \emph{et al.} generate speech from a sequence of silent video frames of a speaking person. Their model uses the video frames and the corresponding optical flow to output a spectrogram representing the speech.
Owens \emph{et al.} \cite{owens2016visually} use a recurrent neural network to predict sound from silent videos of people hitting and scratching objects with a drumstick.

\subsubsection{Audio-visual multi-modal learning}
Recent research in audio-visual speech processing makes extensive use of neural networks. The work of Ngiam \emph{et al.} \cite{ngiam2011multimodal} is a seminal work in this area. They demonstrate cross modality feature learning, and show that better features for one modality (e.g., video) can be learned if both audio and video are present at feature learning time.
Multi-modal neural networks with audio-visual inputs have also been used
for lip reading \cite{chung2016lip}, lip sync \cite{chung2016out} and robust speech recognition \cite{noda2015audio}.

\begin{figure*}[t]
\begin{center}
\includegraphics[width=0.9\linewidth]{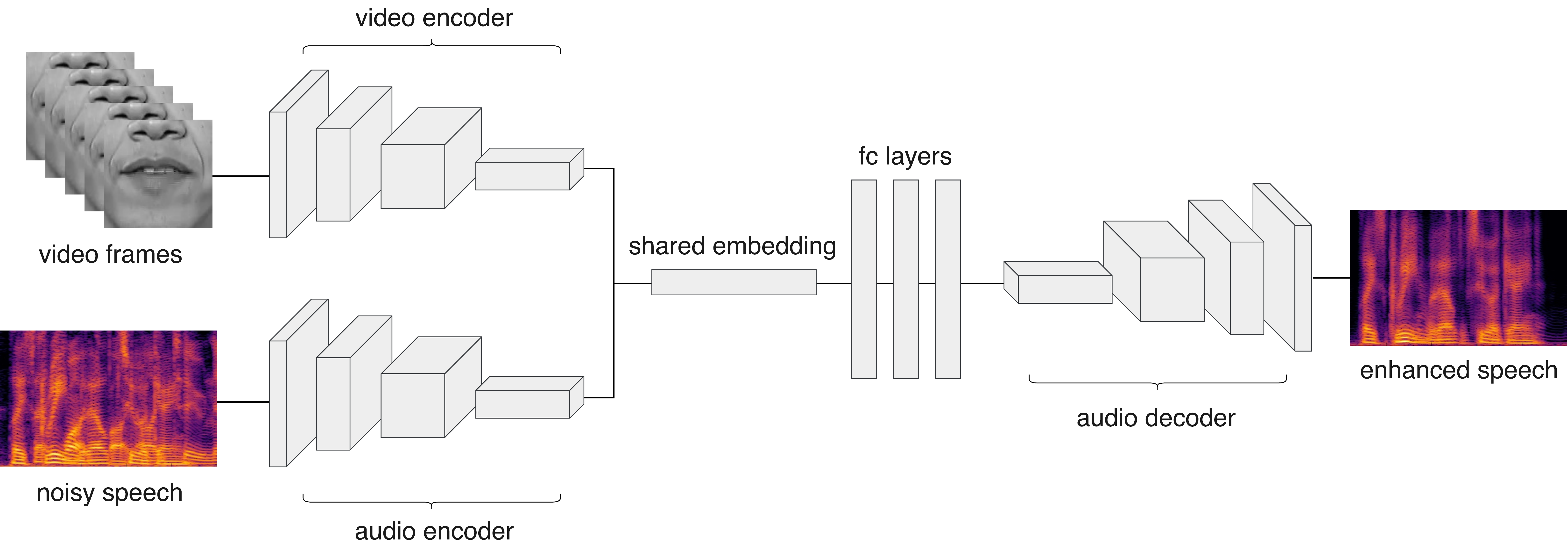}
\end{center}
\caption{Illustration of our encoder-decoder model architecture. A sequence of 5 video frames centered on the mouth region is fed into a convolutional neural network creating a video encoding. The corresponding spectrogram of the noisy speech is encoded in a similar fashion into an audio encoding. A single shared embedding is obtained by concatenating the video and audio encodings, and is fed into 3 consecutive fully-connected layers. Finally, a spectrogram of the enhanced speech is decoded using an audio decoder.}
\label{fig:arch}
\end{figure*}

\subsubsection{Audio-visual speech enhancement}
Work has also been done on audio-visual speech enhancement and separation \cite{Girin2001AudiovisualEO}. Kahn and Milner \cite{khan2013speaker,khan2016audio} use hand-crafted visual features to derive binary and soft masks for speaker separation. Hou \emph{et al.} \cite{hou2017audio} propose convolutional neural network model to enhance noisy speech. Their network gets a sequence of frames cropped to the speaker's lips region and a spectrogram representing the noisy speech, and outputs a spectrogram representing the enhanced speech. Gabbay \emph{et al.} \cite{gabbay2017seeing} feed the video frames into a trained speech generation network \cite{ephrat2017improved}, and use the spectrogram of the predicted speech to construct masks for separating the clean voice from the noisy input.

\section{Neural Network Architecture}
The speech enhancement neural network model gets two inputs: (i) a sequence of video frames showing the mouth of the speaker; and (ii) a spectrogram of the noisy audio. The output is a spectrogram of the enhanced speech. The network layers are stacked in encoder-decoder fashion (Fig.~\ref{fig:arch}).
The encoder module consists of a dual tower Convolutional Neural Network which takes the video and audio inputs and encodes them into a shared embedding representing the audio-visual features. The decoder module consists of transposed convolutional layers and decodes the shared embedding into a spectrogram representing the enhanced speech. The entire model is trained end-to-end.

\subsection{Video encoder}
The input to the video encoder is a sequence of 5 consecutive gray scale video frames of size $128 {\times} 128$, cropped and centered on the mouth region. While using 5 frames worked well, other number of frames might also work. The video encoder has 6 consecutive convolution layers described in Table \ref{tb:encoders}.
Each layer is followed by Batch Normalization, Leaky-ReLU for non-linearity, max pooling, and Dropout of 0.25.

\subsection{Audio encoder}
Both input and output audio are represented by log mel-scale spectrograms having 80 frequency intervals between 0 to 8kHz and 20 temporal steps spanning 200 ms.

As previously done in several audio encoding networks \cite{engel2017neural,grais2017single}, we  design our audio encoder as a convolutional neural network using the spectrogram as input. The network consists of 5 convolution layers as described in Table \ref{tb:encoders}. Each layer is followed by Batch Normalization and Leaky-ReLU for non-linearity. We use strided convolutions instead of max pooling in order to maintain temporal order.

\begin{table}[t]
\centering
\begin{tabular}{|c||c|c||c|c|c|}
	\hline
    \multirow{2}{*}{layer} & \multicolumn{2}{|c||}{video encoder} & \multicolumn{3}{|c|}{audio encoder} \\ \cline{2-6}
     & \# filters & kernel & \# filters & kernel & stride \\ 
    \hline\hline
    1 & 128 & 5 $\times$ 5 & 64 & 5 $\times$ 5 & 2 $\times$ 2 \\ 
    2 & 128 & 5 $\times$ 5 & 64 & 4 $\times$ 4 & 1 $\times$ 1 \\ 
    3 & 256 & 3 $\times$ 3 & 128 & 4 $\times$ 4 & 2 $\times$ 2 \\ 
    4 & 256 & 3 $\times$ 3 & 128 & 2 $\times$ 2 & 2 $\times$ 1 \\ 
    5 & 512 & 3 $\times$ 3 & 128 & 2 $\times$ 2 & 2 $\times$ 1 \\ 
    6 & 512 & 3 $\times$ 3 & & & \\ 
    \hline
\end{tabular}
\bigskip
\caption{Detailed architecture of the video and audio encoders. Pooling size and stride used in the video encoder are always $2 \times 2$, for all six layers.}
\label{tb:encoders}

\end{table}

\subsection{Shared representation}
The video encoder outputs a feature vector having 2,048 values, and the audio encoder outputs a feature vector of 3,200 values. The feature vectors are concatenated into a shared embedding representing the audio-visual features, having 5,248 values. The shared embedding is then fed into a block of 3 consecutive fully-connected layers, of sizes 1,312, 1,312 and 3,200, respectively. The resulting vector is then fed into the audio decoder.

\subsection{Audio decoder}
The audio decoder consists of 5 transposed convolution layers, mirroring the layers of the audio encoder. The last layer is of the same size as the input spectrogram, representing the enhanced speech.

\subsection{Optimization}
The network is trained to minimize the mean square error ($l_2$) loss between the output spectrogram and the target speech spectrogram. We use Adam optimizer with an initial learning rate of $5e^{-4}$ for back propagation.  Learning rate is decreased by 50\% once learning stagnates i.e. the validation error does not improve for 5 epochs.

\section{Multi-Modal Training}
Neural networks with multi-modal inputs can often be dominated by one of the inputs \cite{feichtenhofer2016convolutional}. Different approaches have been considered to overcome this issue in previous work. Ngiam \emph{et al.} \cite{ngiam2011multimodal} proposed to occasionally zero out one of the input modalities (e.g., video) during training, and only have the other input modality (e.g., audio). This idea has been adopted in lip reading \cite{chung2016lip} and speech enhancement \cite{hou2017audio}. In order to enforce using the video features, Hou \emph{et al.} \cite{hou2017audio} adds an auxiliary video output that should resemble the input.

We enforce the exploitation of visual features by introducing a new training strategy. We include in the training data samples where the added noise is the voice of the same speaker. Since separating two overlapping sentences spoken by the same person is impossible using audio only information, the network is forced to exploit the visual features in addition to the audio features. We show that a model trained using this approach generalizes well to different noise types, and is capable to separate target speech from indistinguishable background speech.

\section{Implementation Details}

\subsection{Video pre-processing}
In all our experiments video is resampled to 25 fps. The video is divided to non-overlapping segments of 5 frames (200 ms) each. From every frame we crop a mouth-centered window of size $128 {\times} 128$ pixels, using the 20 mouth landmarks from the 68 facial landmarks suggested by \cite{kazemi2014one}. The video segment used as input is therefore of size $128 {\times} 128 {\times} 5$. We normalize the video inputs over the entire training data by subtracting the mean video frame and dividing by the standard deviation.

\subsection{Audio pre-processing}
The corresponding audio signal is resampled to 16 kHz. Short-Time-Fourier-Transform (STFT) is applied to the waveform signal. The spectrogram (STFT magnitude) is used as input to the neural network, and the phase is kept aside for reconstruction of the enhanced signal. We set the STFT window size to 640 samples, which equals to 40 milliseconds and corresponds to the length of a single video frame. We shift the window by hop length of 160 samples at a time, creating an overlap of 75\%. Log mel-scale spectrogram is computed by multiplying the spectrogram by a mel-spaced filterbank. The log mel-scale spectrogram comprises 80 mel frequencies from 0 to 8000 Hz. We slice the spectrogram to pieces of length 200 milliseconds corresponding to the length of 5 video frames, resulting in spectrograms of size $80 {\times} 20$: 20 temporal samples, each having 80 frequency bins.

\subsection{Audio post-processing}
The speech segments are inferred one by one and concatenated together to create the complete enhanced spectrogram. The waveform is then reconstructed by multiplying the mel-scale spectrogram by the pseudo-inverse of the mel-spaced filterbank, followed by applying the inverse STFT. We use the original phase as of the noisy input signal.

\begin{figure}
\begin{center}
\includegraphics[width=0.45\linewidth]{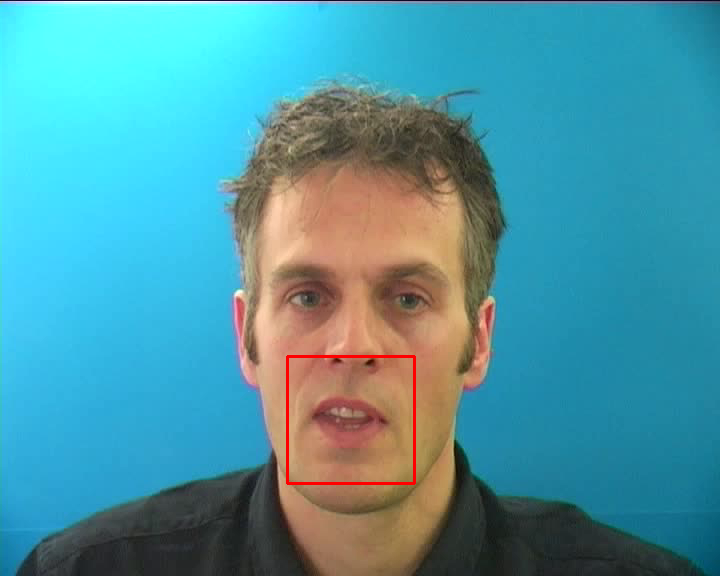}
\includegraphics[width=0.45\linewidth]{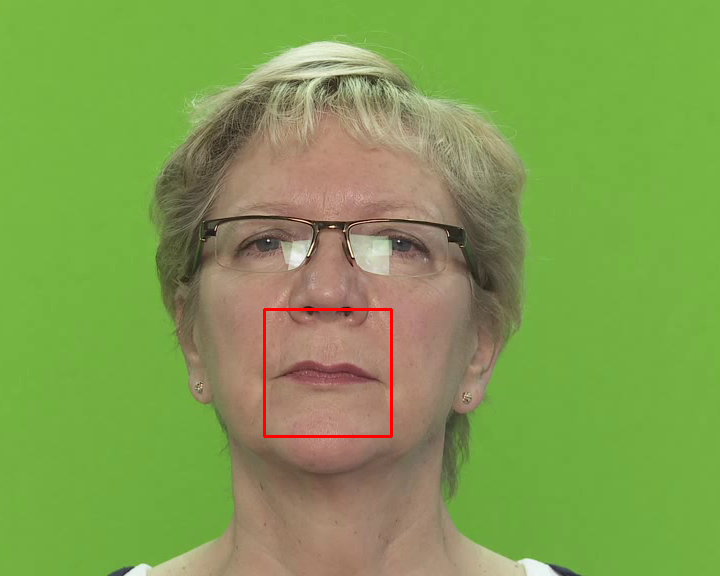}
\includegraphics[width=0.45\linewidth]{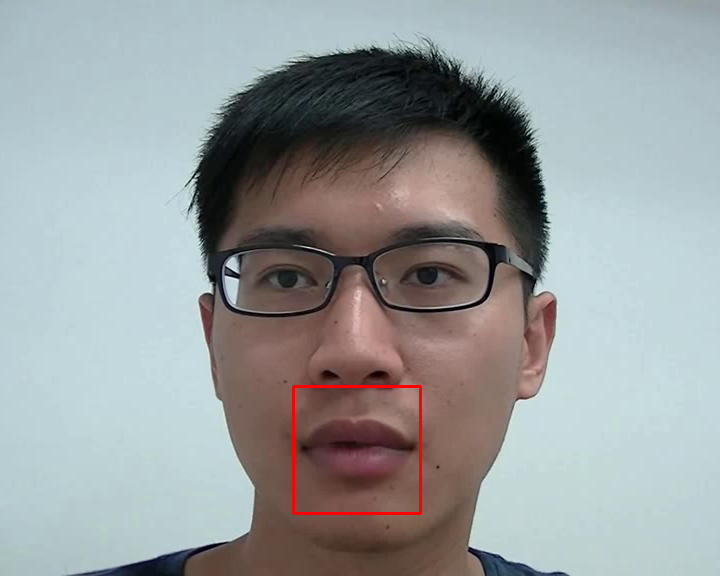}
\includegraphics[width=0.45\linewidth]{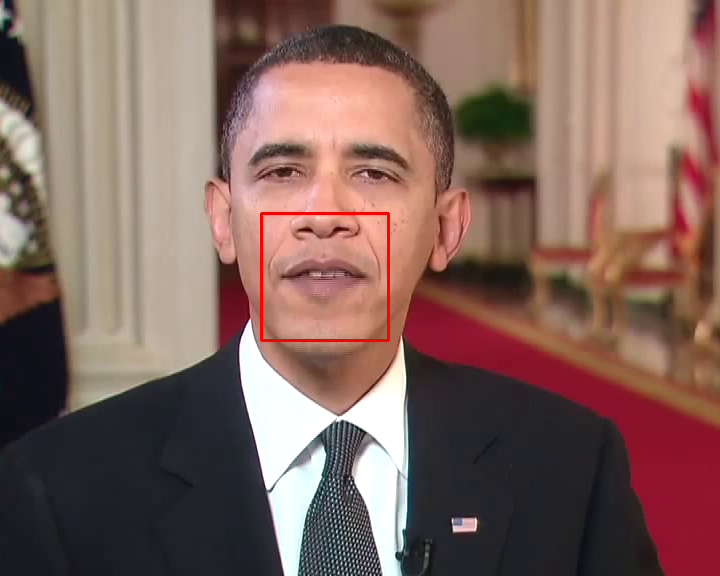}
\end{center}
\caption{Sample frames from GRID (top-left), TCD-TIMIT (top-right), Mandarin speaker (bottom-left) and Obama (bottom-right) datasets. Bounding boxes in red mark the mouth-centered crop region. The Obama videos have varied background, illuminations, resolutions, and lighting.}
\label{fig:datasets}
\vspace{-1em}
\end{figure}

\section{Datasets}
\label{dataset_description}
Sample frames from the datasets are shown in Fig.~\ref{fig:datasets}.

\subsection{GRID Corpus and TCD-TIMIT} 
We perform our experiments on speakers from the GRID audio-visual sentence corpus \cite{gridcorpus}, a large dataset of audio and video (facial) recordings of 1,000 sentences spoken by 34 people (18 male, 16 female). We also perform experiments on the TCD-TIMIT dataset \cite{harte2015tcd} which consists of 60 volunteer speakers with around 200 videos each, as well as three lipspeakers, people specially trained to speak in a way that helps the deaf understand their visual speech. The speakers are recorded saying various sentences from the TIMIT dataset \cite{garofolo1993darpa}.

\subsection{Mandarin Sentences Corpus}
Hou \emph{et al.} \cite{hou2017audio} prepared an audio-visual dataset containing video recordings of 320 utterances of Mandarin sentences spoken by a native speaker. Each sentence contains 10 Chinese characters with phoneme designed to distribute equally. The length of each utterance is approximately 3-4 seconds. The utterances were recorded in a quiet room with sufficient light, and the speaker was captured from frontal view.

\subsection{Obama Weekly Addresses}
We assess our model's performance in more general conditions compared to datasets specifically prepared for lip-reading. For this purpose we use a dataset containing weekly addresses given by Barack Obama. This dataset consists of 300 videos, each of 2-3 minutes long. The dataset varies greatly in scale (zoom), background, lighting and face angle, as well as in audio recording conditions, and includes an unbounded vocabulary.

\section{Experiments}
We evaluate our model on several speech enhancement tasks using the four datasets mentioned in Sec.~\ref{dataset_description}. In all cases, background speech is sampled from the LibriSpeech \cite{panayotov2015librispeech} dataset. For the ambient noise we use different types of noise such as rain, motorcycle engine, basketball bouncing, etc. The speech and noise signals are mixed with SNR of 0 dB both for training and testing, except of the Mandarin experiment where the same protocol of \cite{hou2017audio} is used. In each sample, the target speech is mixed with background speech, ambient noise, or another speech of the target speaker. We call the latter case \emph{self} mixtures.

We report an evaluation using two objective scores: SNR for measuring the noise reduction and PESQ for assessing the improvement in speech quality \cite{pesq}. Since listening to audio samples is essential to understand the effectiveness of speech enhancement methods, supplementary material is available on our project web page \footnote{Examples of speech enhancement can be found at \emph{http://www.vision.huji.ac.il/visual-speech-enhancement}}.

\subsection{Baselines and previous work}
We show the effectiveness of using the visual information by training a competitive audio-only version of our model which has a similar architecture (stripping out the visual stream). Training this baseline does not involve \emph{self} mixtures since audio-only separation in this case is ill-posed. It can be seen that the audio-only baseline consistently achieves lower performance than our model, especially in the \emph{self} mixtures where no improvement in speech quality is obtained at all. In order to validate our assumption stating that using samples of the target speaker as background noise makes the model robust to different noise types as well as cases where the background speech is indistinguishable from the target speech, we train our model once again without \emph{self} mixtures in the training set. It is shown that this model is not capable of separating speech samples of the same voice, although it has access to the visual stream. Detailed results are presented in Table \ref{tb:results}.

We show that our model outperforms the previous work of Vid2speech \cite{ephrat2017improved} and Gabbay \emph{et al.} \cite{gabbay2017seeing} on the two audio-visual datasets of GRID and TCD-TIMIT, as well as achieves significant improvements in SNR and PESQ on the new dataset of Obama. It can be seen that the results are somewhat less convincing on the TCD-TIMIT dataset. One possible explanation might be the smaller amount of clean speech (20 minutes) in the training data compared to other experiments (40-60 minutes). In the Mandarin experiment, we follow the protocol of Hou \emph{et al.} \cite{hou2017audio}, and train our model on their proposed dataset containing speech samples mixed with car engine ambient noise and other interfering noise types, in different SNR configurations. Table \ref{tb:mandarin_results} shows that our model achieves slightly better performance on their proposed test set, while it should be noted that PESQ is not accurate on Chinese \cite{pesq:chinese}.

\begin{table}[tb]
\footnotesize
\renewcommand{\tabcolsep}{2.5pt}
\centering
\begin{tabular}{|l|c|c|c|c|c|c|}
\toprule[1.5pt]
& \multicolumn{3}{c|}{SNR (dB)} & \multicolumn{3}{c|}{PESQ} \\
\midrule
\multicolumn{1}{|c|}{\multirow{2}{*}{Noise type:}} & speech & {amb-} & speech & speech & {amb-} & speech \\
& (other) & ient & (\emph{self}) & (other)& ient & (\emph{self}) \\
\midrule 
\multicolumn{1}{|c|}{\bf{GRID}} & \multicolumn{6}{|c|}{} \\
\midrule
Noisy & 0.07 & 0.26 & 0.05 & 1.91 & 2.08 & 2.15 \\
\emph{Vid2speech} \cite{ephrat2017improved}	& -2.4 & -2.4 & -2.4 & 2.14 & 2.14 & 2.14 \\
\emph{Gabbay} \cite{gabbay2017seeing} & 4.23 & 3.68 & 1.94 & 2.18 & 2.15 & 2.19 \\
Audio-only & 4.61 & 4.43 & 2.03 & 2.57 & 2.58 & 1.96 \\
Ours without \emph{self} & \bf{5.66} & \bf{5.65} & 2.81 & 2.85 & 2.88 & 2.20 \\
Ours with \emph{self} & \bf{5.66} & \bf{5.65} & \bf{4.05} & \bf{2.86} & \bf{2.92} & \bf{2.67} \\
\midrule
\multicolumn{1}{|c|}{\bf{TCD-TIMIT}} & \multicolumn{6}{|c|}{} \\
\midrule
Noisy & 0.01 & 0.03 & 0.01 & 2.09 & 2.28 & 2.21 \\
\emph{Vid2speech} \cite{ephrat2017improved} & -14.25 & -14.25 & -14.25 & 1.27 & 1.27 & 1.27 \\
\emph{Gabbay} \cite{gabbay2017seeing} & 3.88 & 3.84 & 2.62 & 1.71 & 1.81 & 1.82 \\
Audio-only & \bf{5.16} & 5.44 & 1.73 & 2.48 & 2.68 & 1.91 \\
Ours without \emph{self} & 5.12 & \bf{5.46} & 3.28 & \bf{2.62} & \bf{2.70} & \bf{2.22} \\
Ours with \emph{self} & 4.54 & 4.81 & \bf{3.38} & 2.53 & 2.61 & \bf{2.22} \\
\midrule
\multicolumn{1}{|c|}{\bf{Obama}} & \multicolumn{6}{|c|}{} \\
\midrule
Noisy & 0 & 0.01 & 0 & 2.00 & 2.12 & 2.31 \\
Audio-only & 5.06 & 5.7 & 1.84 & 2.44 & 2.56 & 2.11 \\
Ours without \emph{self} & 5.71 & 6.38 & 3.6 & 2.61 & 2.72 & 2.33 \\
Ours with \emph{self} & \bf{6.1} & \bf{6.78} & \bf{5.21} & \bf{2.67} & \bf{2.75} & \bf{2.56} \\
\bottomrule[1.5pt]
\end{tabular}
\bigskip
\caption{Evaluation of our model, with a comparison to baselines and previous work. Our model achieves significant improvement both in noise reduction and speech quality in the different noise types. See text for further discussion.}
\label{tb:results}

\begin{tabular}{|l|c|c|c|c|}
\toprule[1.5pt]
& \multicolumn{2}{c|}{SNR (dB)} & \multicolumn{2}{c|}{PESQ} \\
\midrule
\multicolumn{1}{|c|}{\multirow{2}{*}{Noise type:}} & from \cite{hou2017audio} & speech & from \cite{hou2017audio} & speech \\
&  & (\emph{self}) &  & (\emph{self}) \\ 
\midrule
Input & -3.82 & 0.01 & 2.01 & 2.38 \\
\emph{Hou et al.} \cite{hou2017audio} & 3.7 & - & 2.41 & - \\
Audio-only & 3.23 & 2.09 & 2.30 & 2.24 \\
Ours without \emph{self} & \bf{4.13} & 3.41 & \bf{2.45} & 2.38 \\
Ours with \emph{self} & 3.99 & \bf{4.02} & 2.43 & \bf{2.47} \\
\bottomrule[1.5pt]
\end{tabular}
\bigskip
\caption{Evaluation of our model on the {\bf Mandarin} dataset, along with a comparison to baselines and Hou \emph{et al.} \cite{hou2017audio}, where noise type is speech and ambient.}
\label{tb:mandarin_results}
\vspace{-2em}
\end{table}

\section{Concluding remarks}
An end-to-end neural network model, separating the voice of a visible speaker from background noise, has been presented. Also, an effective training strategy for audio-visual speech enhancement was proposed - using as noise overlapping sentences spoken by the same person. Such training builds a model that is robust to similar vocal characteristics of the target and noise speakers, and makes an effective use of the visual information. 

The proposed model consistently improves the quality and intelligibility of noisy speech, and outperforms previous methods on two public benchmark datasets. Finally, we demonstrated for the first time audio-visual speech enhancement on a general dataset not designed for lipreading research. Our model is compact, and operates on short speech segments, and thus suitable for real-time applications. On average, enhancement of 200 ms segment requires 36 ms of processing (using NVIDIA Tesla M60 GPU).

We note that our method fails when one input modality is missing, since during training both audio and video were used.

The field of combined audio-visual processing is very active. Recent work showing much progress, that appeared just before the camera ready deadline, includes \cite{zissTheConversation}, \cite{ariel}, \cite{berkeley}, \cite{pixels}.

~\\
\noindent
{\bf Acknowledgment.} This research was supported by Israel Science Foundation and Israel Ministry of Science and Technology.

\bibliographystyle{IEEEtran}

\bibliography{mybib}

\end{document}